\documentclass[conference]{IEEEtran}
\IEEEoverridecommandlockouts
\usepackage{cite}
\usepackage{amsmath,amssymb,amsfonts}
\usepackage{algorithmic}
\usepackage{graphicx}
\usepackage{textcomp}
\usepackage{xcolor}
\usepackage{url}
\usepackage{hyperref}
\usepackage[nohyperlinks, nolist]{acronym}

\def\BibTeX{{\rm B\kern-.05em{\sc i\kern-.025em b}\kern-.08em
    T\kern-.1667em\lower.7ex\hbox{E}\kern-.125emX}}

\begin{document}
\title{Self-supervised Optimization of Hand Pose Estimation using Anatomical Features and Iterative Learning \\

\thanks{We thank the Baden-Württemberg Ministry of Economic Affairs, Labour and Tourism for funding the AI Innovation Center ``Learning Systems and Cognitive Robotics" where this work was carried out.}
}

\author{\IEEEauthorblockN{Christian Jauch\IEEEauthorrefmark{1},
Timo Leitritz\IEEEauthorrefmark{1}, and
Marco F. Huber\IEEEauthorrefmark{1}\IEEEauthorrefmark{2}}
\IEEEauthorblockA{\IEEEauthorrefmark{1}Machine Vision and Signal Processing, Fraunhofer Institute for Manufacturing Engineering and Automation IPA, \\Stuttgart, Germany, \texttt{\{firstname.lastname\}@ipa.fraunhofer.de}}
\IEEEauthorblockA{\IEEEauthorrefmark{2}Institute of Industrial Manufacturing and Management IFF, University of Stuttgart, Germany, \texttt{marco.huber@ieee.org}}
}
\maketitle

\begin{abstract}
Manual assembly workers face increasing complexity in their work. Human-centered assistance systems could help, but object recognition as an enabling technology hinders sophisticated human-centered design of these systems. At the same time, activity recognition based on hand poses suffers from poor pose estimation in complex usage scenarios, such as wearing gloves. This paper presents a self-supervised pipeline for adapting hand pose estimation to specific use cases with minimal human interaction. This enables cheap and robust hand pose-based activity recognition. The pipeline consists of a general machine learning model for hand pose estimation trained on a generalized dataset, spatial and temporal filtering to account for anatomical constraints of the hand, and a retraining step to improve the model. Different parameter combinations are evaluated on a publicly available and annotated dataset. The best parameter and model combination is then applied to unlabelled videos from a manual assembly scenario. The effectiveness of the pipeline is demonstrated by training an activity recognition as a downstream task in the manual assembly scenario.
\end{abstract}


\section{Introduction}
Machine vision systems have received a significant boost from machine learning methods, enabling entirely new applications. 
These systems are also making their way into manufacturing environments, where they are enabling new applications to support workers. 
As customers demand more personalized products, the complexity of manual assembly is increasing for workers. At the same time, workers in these areas are often hard to find. Fortunately, new machine learning-based assistance systems can make workers' jobs easier and reduce complexity. 

Today's assistance systems for manual assembly usually rely on object detection. The reason for choosing an object detection-based assistance system for manual assembly is that it is easier to adapt to new processes, thus reducing system maintenance costs. However, usability is not the focus of these systems and a human-centred solution is difficult to implement with object detection as base technology. Furthermore, such systems often interrupt the workers in their work. Activity recognition, on the other hand, can help to avoid interrupting the worker's flow and enable improved human-centred solutions. However, the current state of activity recognition systems limits their utility in an assembly scenario.

There are hand pose-based and image-based activity recognition systems. While image-based systems need a lot of data and thus, are very expensive, pose-based systems focusing solely on the hands of the worker are often not robust enough for use in manual assembly, especially in manufacturing where gloves must be worn. The use of full body pose estimation is often not feasible in manual assembly due to space constraints or lack of detail in the recognition. Neither image-based nor hand pose-based solutions, in their current state, are a viable solution, especially for small and medium-sized enterprises, because there is no model that has been trained on a dataset suitable for the manufacturing domain where it provides the necessary robustness.
Datasets are usually focused on a broad application or have poor data quality, both of which can lead to a drop in accuracy for specific applications. Robustness is particularly important in day-to-day work, as incorrect recognitions unnecessarily reduce productivity, and the assistance system tends to hinder the workers rather than truly assist them.  

One way to enable better human-centred manual assembly assistance systems is to improve pose-based activity recognition to achieve better robustness in the application while keeping the effort required to do so to a minimum. 
The contribution of this paper is a self-supervised pipeline that aims to achieve exactly this. By using anatomical features and confidence thresholds as well as temporal constraints, an application-specific dataset can be generated from unlabeled video input. This dataset is then used to retrain existing models and improve hand pose estimation for the application. Human interaction is kept to a minimum, making the method viable for many different applications where hand pose estimation is beneficial but its accuracy and robustness are not at the required level. The effectiveness of the approach is demonstrated in a downstream task.

The paper is structured as follows: In the next section, existing models and datasets are presented. In Sec.~\ref{sec:method}, the method describing the different steps and parameters of the pipeline is presented. Experimental results are discussed in Sec.~\ref{sec:experiments}. The paper closes with conclusions.

\section{Related Work}
\label{sec:related-work}
\subsection{Manual Assembly Assistance Systems}
Assistance systems for manual assembly typically use object detection methods, e.g.,  the Schlaue Klaus\footnote{https://www.optimum-gmbh.de/produkte/der-schlaue-klaus}, some also include pick-by-light systems such as \cite{2A1_smartAssembly, 2A2_Riedel}. Some systems additionally rely on wearables to track the current progress of the worker \cite{2A3_Sochor}.
Current research explores augmented reality based systems, such as \cite{2A4_MA2RA, 2A5_Tainaka}, using both a head mounted display and projection. For practical reasons and hardware limitations, head mounted displays are not yet widely used in the field, and projection based systems still require visual information from an external camera system. 
This work focuses on camera-based assistance systems with an externally mounted camera. While the integration of e.g. pick-by-light and augmented reality by projection is also possible, the scope of this work is activity recognition. The finalized systems could be used for error detection, training purposes or process documentation. 

\subsection{Hand Pose Estimation}
Hand pose estimation tries to recognise a hand skeleton in an image section containing a hand. For more complex scenes, the image section with a single hand is first selected with the help of an object detector and then the hand skeleton is fitted in this section. 

The object detection task predicts an object, represented by a bounding box with the coordinates of the top left and bottom right corners, combined with a classification of the detected object in an image. In the case of hand pose estimation as a downstream task, object detection predicts the object hand in an RGB image and passes this result to the hand pose estimator.
An in-depth review of object detection can be found in \cite{2B8_Survey}. Current models include single-stage detectors like YOLO \cite{2B2_YOLO} or DETR \cite{2B4_DETR} and two-stage approaches like Faster-RCNN \cite{2B6_FasterRCNN}. Two-stage approaches tend to be slower but more accurate. Recent developments show similar or better accuracy in single stage detectors, while maintaining the speed advantage.

There are several ways to classify existing hand pose estimation algorithms. Firstly, there are different input data for the algorithms, which is also reflected in the available datasets. While at the beginning of the machine learning era the focus lay on depth images 
\cite{2C2_DeepPrior}, their limitations were quickly recognised and RBG data or combined RGB-D data were used as input \cite{2C3_Panteleris, 2C14_DARK}.
Currently, research is mainly focused on RGB data, which is also very practical for applications due to the low hardware requirements. 

Second, current algorithms also differ in their output. Classically, hand pose estimators output a hand skeleton consisting of a varying number of joints---typically 21 joints are specified, 4 per finger and an additional one for the wrist. 
The joints are typically called keypoints. Another output option is a hand mesh \cite{2C6_Mesh} based on hand models such as MANO \cite{2C7_EmbodiedHands}. The 21 joints can also be derived from this, but the mesh is not necessary for the application considered in this work. Therefore, only methods that estimate the joint points directly from an RGB image are considered. Each joint can be represented as two-dimensional pixel coordinate or as three dimensional point in space.  Since activity recognition usually works better with two-dimensional poses due to improved data quality and hand pose estimation in manual assembly has a lot of occlusion, three-dimensional estimators are neglected \cite{2C8_skeletonaction}. 

There is also a large overlap between hand pose and human pose estimation and often the same models can be used with minor adjustments \cite{2C11_WholeBody}.
Current state of the art models are presented in \cite{2C12_HighRes, 2C13_Hourglass, 2C14_DARK}, the latter being an extension for other models. A comprehensive review of hand pose estimation can be found online\footnote{\label{foot:awesome-hand-pose}Xinghaochen on GitHub: \url{https://github.com/xinghaochen/awesome-hand-pose-estimation}}. 

\begin{figure*}[tb]
\centerline{\includegraphics[width=\textwidth]{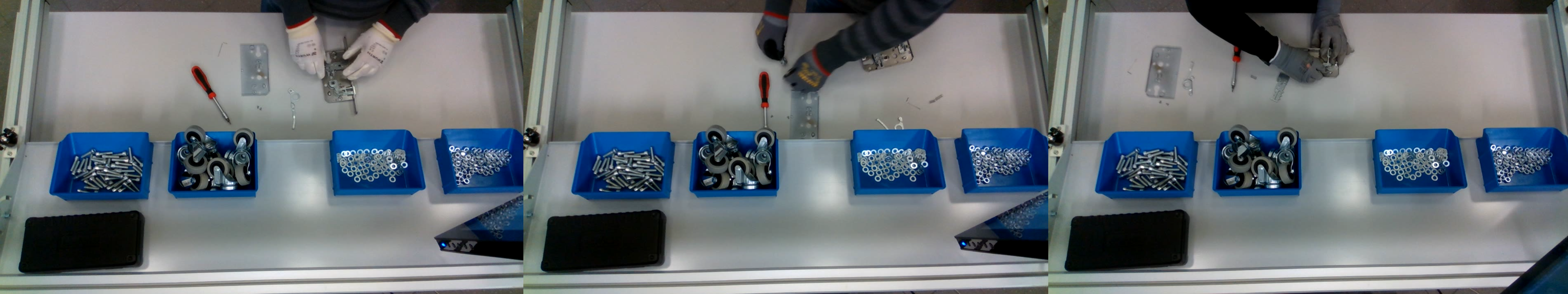}}
\caption{Examples of the manual assembly dataset recorded at Fraunhofer IPA. A simple lock is assembled by three different people all of them wearing three different types of gloves and no gloves. Each person assembles the lock twice with each glove and bare hands.}
\label{fig:dataset}
\end{figure*}

\subsection{Hand Pose Datasets}\label{ch:2c}
The hand pose estimation datasets are also listed in\footref{foot:awesome-hand-pose}. The datasets have a great influence on how well the model will perform in the real application. Datasets with RGB data as input and 21 joints as output are available from different perspectives, with the 3rd person view being suitable for manual assembly. Considering these criteria, possible datasets are Interhand2.6M \cite{2D1_Interhand}, OneHand10K \cite{2D3_OneHand}, FreiHand \cite{2D4_Freihand} with its extension HanCo \cite{2D5_Zimmermann}, RHD2D \cite{2C4_Zimmermann}, and COCOHand \cite{2D6_COCOHand}. Considering whole body pose estimators, the COCO WholeBody dataset \cite{2C11_WholeBody} may also be suitable for evaluation. 

Given the application, the initial models used for evaluation are trained on the OneHand10K, RHD2D, COCOHand, FreiHand and COCOWholeBody datasets. As the pipeline considers temporal information, a dataset with temporal information is required for the evaluation. Also, real videos will be fed to the pipeline, but no synthetic data will be generated, as the pipeline does not include a simulation environment. HandCo provides both real data and temporal context and is the best dataset found for the evaluation. 
The Assembly101 dataset \cite{2D7_Assembly101} is very close to the application, but at the time of the paper camera calibrations of the pose data for 3rd person view were not available.
\label{datasetIPA}
In addition, an application-specific dataset is recorded in the Vision Lab of Fraunhofer IPA. Here, multiple people are assembling a lock while wearing different gloves or no gloves. The test data is separated by person and glove type to ensure that there is no test data leaking into the training set.  Examples are given in Fig.~ \ref{fig:dataset}.

\begin{figure}[tb]
\centerline{\includegraphics[width=0.5\textwidth]{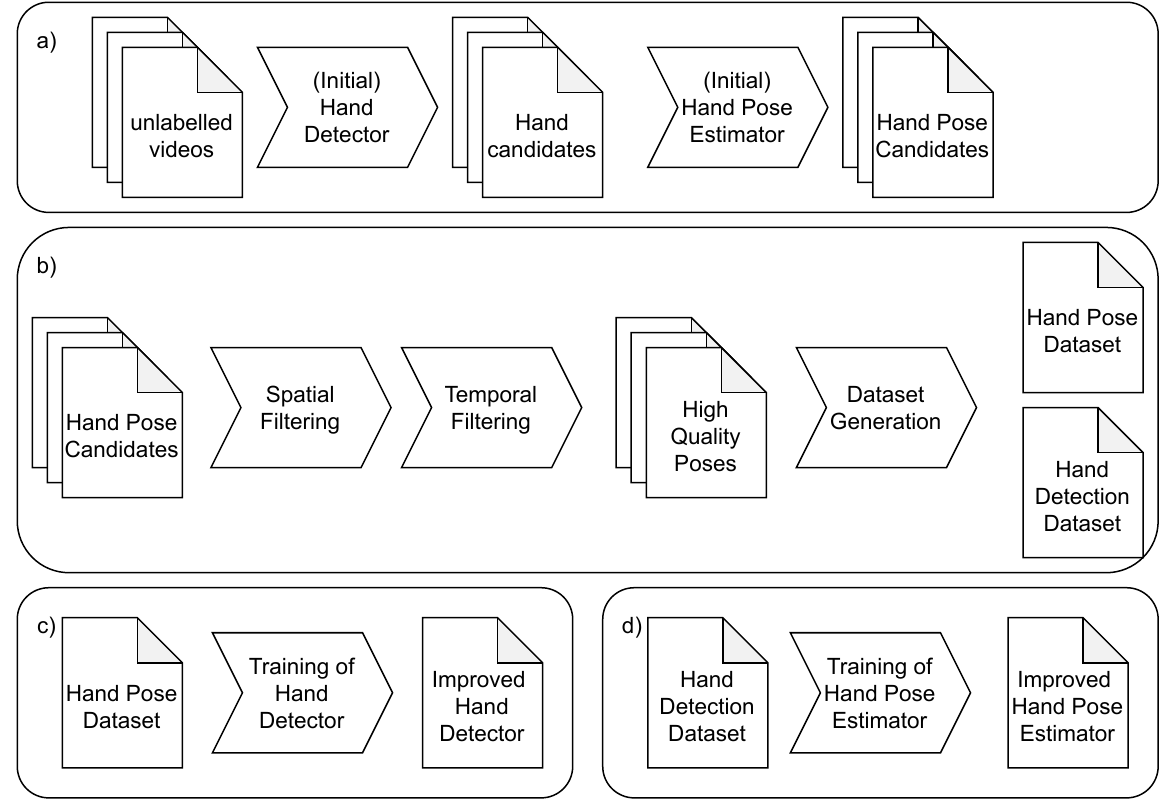}}
\caption{Overview of the  pipeline. a) Initial generation of hand pose dataset candidates. b) Filtering of candidates and generation of dataset. c) Retraining of hand detector with dataset from step b). d) Retraining of pose estimator with dataset from step b). After performing step c) and d), step a) and b) is repeated with the updated models.}
\label{fig:overview}
\end{figure}

\section{Method}
\label{sec:method}
An overview of the pipeline is provided, including three key components that are described in more detail.
\subsection{Overview}
Fig.~\ref{fig:overview} shows an overview of the pipeline. In order to enable a fast, cheap and simple adaptation of the machine learning models to the application, manual annotation of the data is not possible. Therefore, the presented method relies on unlabeled videos from the specific application as input data. 

In a first step, shown in Fig.~\ref{fig:overview}\,a), possible hands in the image are first identified with the help of an object detector or, more precisely, a hand detector. For the initial iteration, a hand detector is used that has been trained on a publicly available, generalized hand dataset. The hand detector returns bounding boxes containing the image regions of possible hands. These image regions are then fed into a hand pose estimator, which is also trained on a publicly available general dataset in the first iteration. 

The result of step a) is, in addition to the bounding boxes, the keypoints of the possible hand poses. For each previously detected bounding box, an attempt is made to fit a hand pose, regardless of whether it exists or not. These results can be used to identify wrong detections in the first place by removing low scoring pose estimates and their corresponding hand detections. In the next step b), the hand pose candidates are filtered using spatial and temporal information. Spatial filtering uses anatomical information (i.e., expected maximum bone length, minimum and maximum hand size) and contextual knowledge (maximum number of hands present in the video at the same time) to remove bad candidates. Temporal filtering is then used to detect and smooth jitter in the keypoints and interpolate between frames. Poor initial estimates are removed from the pipeline, leaving a smaller dataset with improved data quality (lower recall, higher precision). Two datasets are created from the poses: one for retraining the hand detector and one for retraining the hand pose estimator.
Each dataset is then used to retrain an updated model for hand detection in step c) and hand pose estimation in step d). The process is then repeated iteratively with the updated hand detection model and hand pose estimation model in step a). The optimization stops after a user-defined number of iterations. In the experiments it was found that already three iterations provide satisfactory results. 

\subsection{Initial Model Candidates}
The model training is performed with the MMLab framework \cite{3A1_MMLab}. 
Initially, a total of five model combinations are evaluated. Four out of five models use a hand detector and hand pose estimator. For hand detection a pre-trained Cascade RCNN provided by MMDetection is used. For hand pose estimation, all three models but one are based on the same architecture, which is an HRNet. These models merely differ on the dataset used for training the HRNet. For this purpose, the datasets mentioned in Sec. ~\ref{ch:2c} are chosen. The forth hand pose model acts as a baseline. It is built on the SimpleBaseline model \cite{3A2_SimpleBaseline} and is pre-trained on the FreiHand dataset. This dataset is similar to the HanCo dataset in terms of hand angle and size, but has very different backgrounds. This may be an advantage for this model in the first step.

The whole body pose estimation model based on the COCOWholeBody dataset is also using an HRNet architectur. The object detection focuses on people instead of hands and is a pre-trained Faster-RCNN.

\subsection{Spatial Filtering}
The spatial filtering pipeline consists of three different checks using contextual knowledge and anatomical features. First, each bone is checked for its maximum length. The bone length is only checked as an upper bound, as the lower bound cannot be determined in a two-dimensional case. The maximum length of the first index finger bone from the palm in pixels is the parameter $s_\mathrm{bone}$ and depends on the camera and its placement. The results of \cite{3B1_Bones} are used to calculate the maximum length of each bone depending on the length provided. If one bone exceeds the maximum length, the hand is removed  from the dataset as a wrong detection.

Next, the size of the hand is checked. Depending on the distance between the hand and the camera, only a certain percentage $s_\mathrm{Area,max}$  of the image can be filled with the hand. 
Also, a certain percentage of the image $s_\mathrm{Area,min}$ must be filled with the hand. If the hand area, the bounding box of the hand detector, exceeds the maximum area or is less than the minimum area, the hand is again removed from the dataset as a wrong detection.

If there is not the expected amount of hands $s_\mathrm{count}$ that the contextual knowledge suggests in the image, the filter removes the excess hands with the lowest confidence score or removes the whole image from the dataset. Accepting fewer hands than expected resulted in bad hand detection performance.

\subsection{Temporal Filtering}
The temporal filter is responsible for removing jumps between images and interpolating bad keypoints and missing detections. 
Jumping detections are an indication of incorrect detections between images. This is done by specifying a maximum velocity of a single keypoint or bounding box corner $t_\mathrm{vmax}$. The velocity of each point is then checked, and if the maximum velocity is exceeded, the joint or detection itself (if the point is part of the bounding box) is removed. 
Afterwards, the filter attempts to linearly interpolate each point between up to five individual images, if possible.

\begin{figure}[tbp]\centerline{\includegraphics[width=0.5\textwidth]{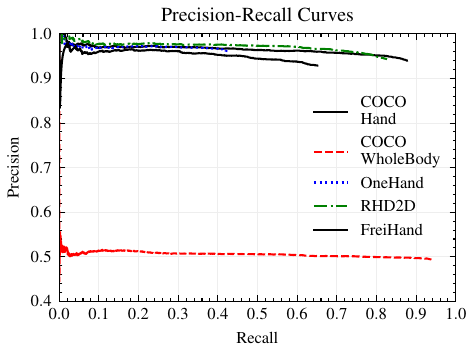}}
\caption{Precision and recall curves of the initial model candidates.}
\label{fig:prcurve}
\end{figure}

\begin{figure}[tbp]
\centerline{\includegraphics[width=0.5\textwidth]{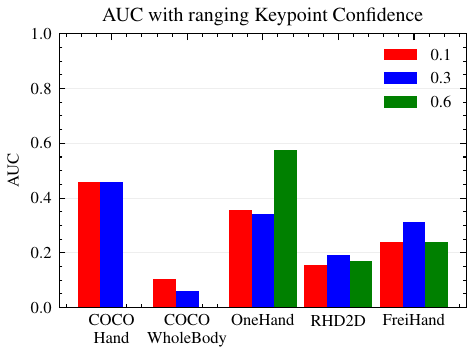}}
\caption{AUC over different keypoint thresholds of the initial model candidates.}
\label{fig:AUCinitial}
\end{figure}

\section{Experimental Results}
\label{sec:experiments}
This section presents the experimental results and includes a description of the evaluation datasets. Two types of experiments are performed in different configurations. Experiment~\#1 validates different aspects of the pipeline by comparing them to an available ground truth. Experiment~\#2 builds on the results of the first experiments and evaluates the entire pipeline. The parameters for both experiments are shown in Table~\ref{tab2:Parameters}.

\subsection{Dataset}
Since the video data for the manual assembly task is unlabelled, another dataset is used to evaluate different configurations of the pipeline and show its effectiveness, using 100 randomly sampled videos with a random homogenization background from the HanCo dataset \cite{2D5_Zimmermann}. After finding the pipeline configuration, it is then applied to the unlabelled data from the manual assembly task in Section~\ref{datasetIPA}. To evaluate the performance of the pipeline in this scenario, a test video is processed by the initial models without the proposed pipeline and by the models after retraining on multiple iterations with the proposed pipeline.

\subsection{Initial Model Selection}
The initial model is selected by comparing the precision and recall curves of the hand detection models with a small confidence threshold for the hand detection as well as the pose estimation. Also, the \ac{AUC} metric of the pose estimators are considered in the selection process. 

Fig.~\ref{fig:prcurve} shows the precision-recall curves for the five dataset models with a confidence threshold for pose estimation of $c_\mathrm{pe}=0.1$. Except for COCOWholeBody, all models show high precision values from the start, while RHD2D and COCOHand also reach high recall values while maintaining precision. Considering the data, this is to be expected as HanCo only includes hand areas. In an application where the whole body is visible, the use of COCOWholeBody as the initial model might be more appropriate. 

Fig.~\ref{fig:AUCinitial} shows the \ac{AUC} at different confidence levels for keypoint estimation. COCOHand and OneHand both show promising results with an \ac{AUC} above 0.45 for COCOHand for $c_\mathrm{pe} = 0.1$ and $c_\mathrm{pe} = 0.3$. The \ac{AUC} of OneHand is 0.34 for $c_\mathrm{pe} = 0.1$ and $c_\mathrm{pe} = 0.3$ and the \ac{AUC} is larger than 0.55 for $c_\mathrm{pe} = 0.6$. COCOHand is not confident enough to estimate any pose when $c_\mathrm{pe}$ is set to 0.6 or higher. However, according to the precision-recall curves, there are fewer candidates for OneHand due to the lower recall, so COCOHand is selected as the first candidate model. 

\begin{figure}[tbp]
\centerline{\includegraphics[width=0.5\textwidth]{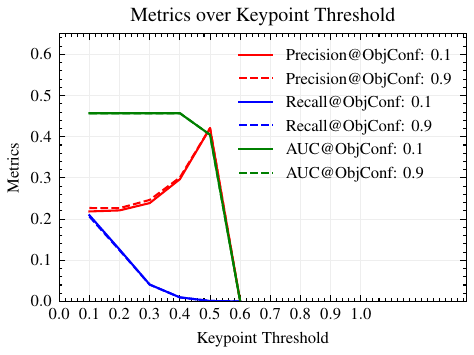}}
\caption{Evaluation of different Confidence Thresholds with IoU=0.75.}
\label{fig:confids}
\end{figure}

To find the optimal confidence thresholds for the initial model, precision, recall and \ac{AUC} are evaluated for different object confidence thresholds $c_\mathrm{hd}$ and different keypoint thresholds $c_\mathrm{pe}$. Fig.~\ref{fig:confids} shows the results. $c_\mathrm{hd}$ has little effect on the data. The \ac{AUC} is stable for all values below 0.5, while precision and recall vary. Different values can be used depending on the focus. The authors decided that the improved recall of 0.2 was more valuable than the lost precision. Therefore, the thresholds are fixed at $c_\mathrm{hd}=0.9$ and $c_\mathrm{pe}=0.2$ for further testing.

\begin{table}[tbp]
\caption{Parameters}
\label{tab2:Parameters}
\begin{center}
\begin{tabular}{|c|c|c|c|c|c|}
\hline
Dataset & $s_\mathrm{bone}$& $s_\mathrm{Area,max}$ & $s_\mathrm{Area,min}$& $s_\mathrm{count}$& $t_\mathrm{vmax}$ \\
\hline
HanCo & $50px$ &  $75\%$ & $15\%$ & $1$ & $25px$\\
Assembly & $80px$ &  $80\%$ & $5\%$ & $2$ & $45px$\\
\hline
\end{tabular}
\end{center}
\end{table}

\begin{table}[tbp]
\caption{Metric table of tests on HanCo dataset.}
\label{tab1:Metrics}
\begin{center}
\begin{tabular}{|c|c|c|c|c|c|c|}
\hline
\textbf{}& &\multicolumn{2}{|c|}{\textbf{Metrics@0.5}}&\multicolumn{2}{|c|}{\textbf{Metrics@0.75}}& \\
\cline{3-7} 
\textbf{\#} & Config & \textbf{\textit{Precision}}& \textbf{\textit{Recall}} & \textbf{\textit{Precision}}& \textbf{\textit{Recall}}& \textbf{\textit{AUC}} \\
\hline
1&No Filter&  0.9472 & 0.5198 & 0.2266 & 0.1244 & 0.4571 \\
1&Spatial& 0.9858 & 0.0212 & 0.4184 & 0.0090 & 0.3250 \\
1&Both& 0.9479 & 0.5196 & 0.2269 & 0.1244 & 0.4345 \\ \hline
2&Initial& 0.9482 & 0.4663 & 0.2314 & 0.1138 & 0.1952 \\ 
2&Retrained& 0.9469 & 0.6670 & 0.2540 & 0.1789 & 0.2857 \\ 
\hline
\end{tabular}
\end{center}
\end{table}

\subsection{Effects of Spatial and Temporal Filtering}
Table~\ref{tab1:Metrics} shows the effect of the spatial and temporal filters in Experiment \#1. For comparison, the metrics are shown without the filter, with the spatial filter only, and with both filters. Because the hand is always in the center of the image and almost fills the whole image, the metrics are very good with an IoU of 0.5, as expected. With an accepted IoU of 0.75, the values drop sharply. The spatial filter significantly reduces the recall while increasing the precision. The effect of the temporal filter is exactly the opposite, reaching values similar to those without any filter. The temporal filter adds new detections with linear interpolation. Given the comparatively low frame rate of the HanCo dataset, this does not work very well. In applications with higher frame rates the approach will be more precise, as will be shown below.

\begin{table}[tbp]
\caption{Tests on manual assembly dataset.}
\label{tab2:Metrics}
\begin{center}
\begin{tabular}{|c|c|c|c|}
\hline
\textbf{Config} & \textbf{\textit{\# of frames}} & \textbf{\textit{\# with correct}}& \textbf{\textit{\# with correct}} \\
 & \textbf{\textit{with hands}} & \textbf{\textit{hands}}& \textbf{\textit{poses}} \\
\hline
Initial model& 2633 & 0 & 171 \\
Retrained model&  2633 & 1740 & 2211  \\
\hline
\end{tabular}
\end{center}
\end{table}

\subsection{Comparison with and without Retraining on Test data}
Experiment \#2 in Table~\ref{tab1:Metrics} shows the effect of retraining on the HanCo dataset. 20 randomly chosen videos, which were not part of the retraining, are used for evaluation. The retraining was performed for three iterations and the evaluation was done without the proposed filters using $c_\mathrm{hd}=0.9$ and $c_\mathrm{pe}=0.2$. The initial values are again very good due to the simplicity of the used dataset. However, an improvement can still be seen due to the retraining, especially in the recall.

Table~\ref{tab2:Metrics} shows the effect of the proposed pipeline for the manual assembly application from section \ref{datasetIPA}. The evaluation is performed manually frame by frame on a test video where the worker wears gloves. For each image it is decided whether all objects and all poses of the recognized objects have been recognized satisfactorily. A recognition is considered correct if all hands are correctly recognised. Poses can be considered correct even if not all objects are correctly recognised. 
The first configuration uses the initial untrained models, while the second uses the retrained models after three iterations. All confidence thresholds are set to $c_\mathrm{hd}=c_\mathrm{pe}=0.1$. The performance of the retrained model significantly improves detection and estimation, even when the worker is wearing gloves and detects about $66\%$ of all hands. The biggest challenge was motion blur, which resulted in only one hand being detected and the frame being counted as a false positive.

\subsection{Downstream Task Activity Recognition}
The downstream task is tested with the PoseConv3D model \cite{2C8_skeletonaction} using the generated poses after retraining on the manual assembly dataset described in Section~\ref{datasetIPA}. PoseConv3D was trained with a batch size of 8 for 100 epochs with a learning rate of 0.0002 using stochastic gradient descent and a clip size of 5 while applying the uniform sampling technique that was proposed in \cite{2C8_skeletonaction}. 

Five activity classes are evaluated: reach, grasp, bring, join, and release. The test is only performed after retraining, as the baseline model completely fails to generate accurate poses due to the use of gloves in the dataset. After training, the activity recognition model achieves a top1 accuracy of $0.750\pm0.016$ and a mean class accuracy of $0.738\pm0.022$ averaged over three runs. In conclusion, activity recognition would not have been possible at all if the worker had worn gloves with the baseline models, but was enabled by the proposed pipeline.

\section{Conclusions}
\label{sec:conclusion}
The proposed pipeline was evaluated on a subset of the HanCo dataset and different initial models and parameters were evaluated to find a suitable pipeline for demanding applications. The results were promising on the HanCo dataset and iterative training improved the results even further. The pipeline exploits anatomical features in combination with a fixed camera configuration and, coupled with iterative training, the models are adapted to the application. The pipeline was also evaluated on an unlabelled manual assembly dataset recorded at the Fraunhofer IPA as a challenging application involving not only bare hands but also different gloves. The results are also shown in an activity recognition task, which was not possible with the original models. The proposed work is a first step towards changing the way workers interact with manual assembly assistance systems in the future. 

\bibliographystyle{IEEEtran}
\bibliography{IEEEabrv,conference_101719}

\begin{acronym}
\acro{AUC}[AUC]{Area under Curve}
\end{acronym}

\end{document}